\documentclass[sigconf,nonacm]{acmart}

\setcopyright{rightsretained}
\copyrightyear{2023}
\acmYear{2023}
\acmConference[SACMAT '23]{Proceedings of the 28th ACM Symposium on Access Control Models and Technologies}{June 7--9, 2023}{Trento, Italy}
\acmBooktitle{Proceedings of the 28th ACM Symposium on Access Control Models and Technologies (SACMAT '23), June 7--9, 2023, Trento, Italy}
\acmDOI{10.1145/3589608.3595078}
\acmISBN{979-8-4007-0173-3/23/06}

\usepackage[skip=0pt]{subcaption}

\usepackage[capitalize]{cleveref}
\crefname{lstlisting}{Listing}{Listings}

\usepackage{listings}
\usepackage{color}

\lstdefinelanguage{PDDL}
{
  sensitive=false,    
  morecomment=[l]{;}, 
  alsoletter={:,-},   
  morekeywords={
    define,domain,problem,not,and,or,when,forall,exists,either,
    :domain,:requirements,:types,:objects,:constants,
    :predicates,:action,:parameters,:precondition,:effect,
    :fluents,:primary-effect,:side-effect,:init,:goal,
    :strips,:adl,:equality,:typing,:conditional-effects,
    :negative-preconditions,:disjunctive-preconditions,
    :existential-preconditions,:universal-preconditions,:quantified-preconditions,
    :functions,assign,increase,decrease,scale-up,scale-down,
    :metric,minimize,maximize,
    :durative-actions,:duration-inequalities,:continuous-effects,
    :durative-action,:duration,:condition
  }
}

\begin{document}
\title[How to Raise a Robot --- A Case for Neuro-Symbolic AI in Constrained Task Planning]{How to Raise a Robot --- A Case for Neuro-Symbolic AI in Constrained Task Planning for Humanoid Assistive Robots}

\author{Niklas Hemken}
\email{niklas.hemken@student.kit.edu}
\orcid{0009-0001-3620-3462}
\affiliation{%
  \institution{KIT}
  \streetaddress{Kaiserstraße 12}
  \city{Karlsruhe}
  \country{Germany}
 \postcode{76131}
}
\author{Florian Jacob}
\email{florian.jacob@kit.edu}
\orcid{0000-0002-5739-8852}
\affiliation{%
  \institution{KIT}
  \streetaddress{Kaiserstraße 12}
  \city{Karlsruhe}
  \country{Germany}
 \postcode{76131}
}
\author{Fabian Peller-Konrad}
\email{fabian.peller-konrad@kit.edu}
\orcid{0000-0001-8120-933X}
\affiliation{%
  \institution{KIT}
  \streetaddress{Kaiserstraße 12}
  \city{Karlsruhe}
  \country{Germany}
  \postcode{76131}
}
\author{Rainer Kartmann}
\email{rainer.kartmann@kit.edu}
\orcid{0000-0002-8891-9366}
\affiliation{%
  \institution{KIT}
  \streetaddress{Kaiserstraße 12}
  \city{Karlsruhe}
  \country{Germany}
  \postcode{76131}
}
\author{Tamim Asfour}
\email{asfour@kit.edu}
\orcid{0000-0003-4879-7680}
\affiliation{%
  \institution{KIT}
  \streetaddress{Kaiserstraße 12}
  \city{Karlsruhe}
  \country{Germany}
  \postcode{76131}
}
\author{Hannes Hartenstein}
\email{hannes.hartenstein@kit.edu}
\orcid{0000-0003-3441-3180}
\affiliation{%
  \institution{KIT}
  \streetaddress{Kaiserstraße 12}
  \city{Karlsruhe}
  \country{Germany}
  \postcode{76131}
}

\renewcommand{\shortauthors}{Niklas Hemken et al.}

\begin{abstract}
Humanoid robots will be able to assist humans in their daily life, in particular due to their versatile action capabilities.
However, while these robots need a certain degree of autonomy to learn and explore, they also should respect various constraints, for access control and beyond.
We explore the novel field of incorporating privacy, security, and access control constraints
with robot task planning approaches.
We report preliminary results on the classical symbolic approach, deep-learned neural networks, and modern ideas using large language models as knowledge base.
From analyzing their trade-offs, we conclude that a hybrid approach is necessary, and thereby present a new use case for the emerging field of neuro-symbolic artificial intelligence.
\end{abstract}

\ccsdesc[500]{Security and privacy~Access control}
\ccsdesc[500]{Computer systems organization~Robotics}
\ccsdesc[500]{Computing methodologies~Neural networks}
\ccsdesc[500]{Computing methodologies~Robotic planning}

\keywords{Humanoid Robots, Robot Task Planning, Activity-Centric Access Control, Deep Learning based Access Control, Large Language Models, Neuro-Symbolic Access Control}

\settopmatter{printfolios=true}
\maketitle

\section{Introduction}

\begin{figure}
    \centering
    \includegraphics[width=\linewidth, trim=1.6cm 3.6cm 2.0cm 3.3cm, clip]{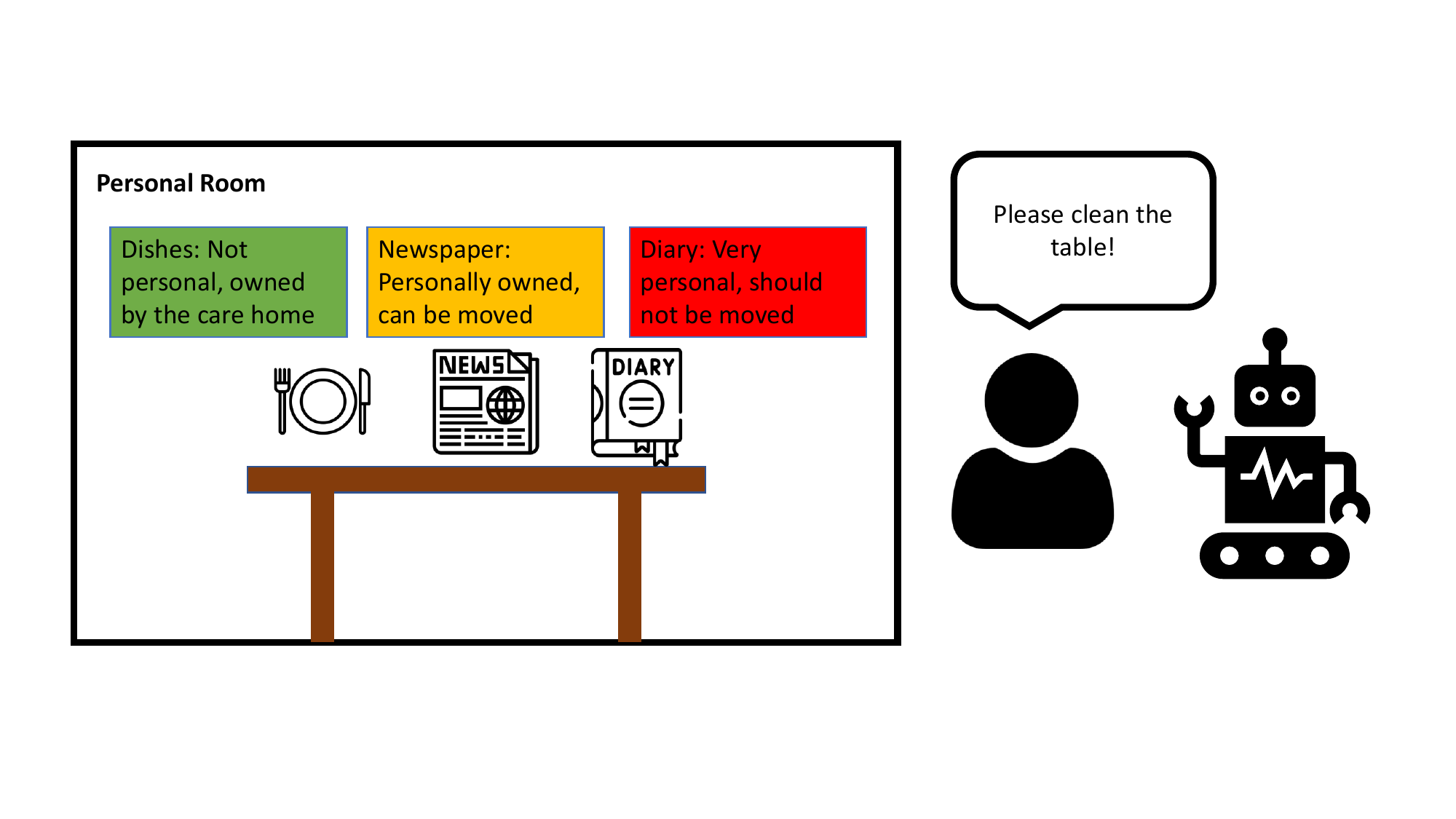}
    \caption{An assistive robot is tasked to clean the table in the personal room of a care home resident.
    The challenge is to integrate an abstract “do not move private objects” policy as specific constraints into robot task planning.
    While privacy has a common understanding, it is a subjective preference that needs ad-hoc assessment by the robot.
    On the table are care home-owned dishes, thus not personal.
    The newspaper is personally-owned but contains no private information.
    The diary is personal and private.
    }
    \label{fig:example}
\end{figure}

With an intensifying labor shortage in the care sector, assistive humanoid robots will most likely become a necessity in the future of our aging society.
These robots will need to invade the most private spaces of humans they care for, thus safety, security and privacy issues in this field are of utmost importance.
The potential of humanoid robots to assist humans lies in the ability to learn whatever is needed for assistance.
However, `whatever is needed' has to be restricted to safeguard safety, security, and privacy policies and preferences.
The challenge for classical approaches, i.e., based on symbols and logical formulas, to ensure constraints lies in the task universality of humanoid robots.
Their wide range of tasks and deployment favors approaches in which the robot is granted increased sovereignty and should learn its authorizations in the field.
Learning promises scalability of problem complexity up to task universality, while keeping manual specification complexity manageable for humans.
However, learning changes the nature of access control from certain enforcement to probable observance.
Future use cases of assistive robots in care home will require complex systems of requirements regarding safety, security and a lot of personalized requirements when directly interacting with patients.
While some of these requirements are prescribed in policy documents like the official instructions for human care-home workers accessible to the robot via manual translation or machine learning, some requirements are purely based on “common sense” and a common understanding of social norms not compiled in a machine-readable way.
An example of such norm is what objects a human considers as private belongings and what objects can inoffensively be interacted with.
In this tech report, we investigate how symbolic and neural methods to robot task planning can ensure symbolic and neural constraints beyond access control, and how constraints from policy documents and social norms can be inferred and observed using large language models.
We discuss the inherent trade-offs, and propose neuro-symbolic hybrid methods as vision to bring assisted living with humanoid robots into practice.

As toy example illustrated in \cref{fig:example}, a robot is tasked to clean a table in a personal room, for which the robot now needs a plan.
It considers whether it is allowed to move objects on the table, i.e., the 
authorization to move an object is a precondition of the action for moving that object.
We assume constraints that formalize policies like  ``do not move personal items'', which allow, e.g., the removal of care home owned dishes, but disallow moving a diary.
To put user-instructed policies like that in concrete terms, the robot needs a kind of policy information database or knowledge base to decide whether an action with an object is actually constrained by a policy.

\section{Related Work}

In a first attempt to tackle this problem of balancing sovereignty and `obedience', we proposed a joint modeling framework for robot task planning and access control~\cite{bayreuther2022bluesky}.
By directly incorporating access control into task planning, robots are unable to even “think about” forbidden behavior.
This tech report is an extended version of our poster abstract~\cite{how-to-raise-a-robot-poster}.
While the fit between the notion of an activity in Activity-Centric Access Control (ACAC)~\cite{gupta2021towards} and task planning was shown in theory, we now report on preliminary practical insights.
Recently, machine-learning-based approaches to assist or even completely perform access decisions have been proposed, as seen with Machine-Learning- and Deep-Learning-based Access Control (MLBAC/DLBAC)~\cite{nobi2022administration}.

\section{Problem Statement and Definitions}
\label{sec:bg}

We describe the problem as follows.
A prompt is an instruction given by a user to perform a task to reach a certain goal.
A signal-based prompt is, e.g., a user-written text or a microphone recording.
A symbol-based prompt is formalized and refers to abstract symbols, like subjects, objects, and locations with attributes.
A plan is a sequence of actions to be performed by the robot to fulfill the task and reach the goal based on its current initial state.
A symbol-based plan is a sequence of abstract actions with their pre- and postconditions (effects) that is converted to a signal-based plan in the form of actuator commands needed for execution.
Planning is the problem of finding such a sequence of actions.
Symbolic planning does so by reshaping logical formulas,
and needs a symbol-based prompt as input to output a symbolic plan.
Symbolic planning can be implemented either with classical search algorithms or symbolic artificial intelligence, but always uses a localist data representation where one symbol is represented in one discrete variable in information processing.
Neural planning is the forward pass through a deep neural network that learned to plan, and can either use a symbol- or signal-based prompt, to either create a symbol- or signal-based plan.
End-to-end planning infers a signal-based plan from a signal-based prompt.
Neural information processing usually operates on a distributed data representation where one specific representation of a symbol in the respective feature space is mapped to not one single neuron, but a continuous activation pattern of many neurons.
The distributed data representation is the key that allowed deep neural networks to advance past symbolic methods in terms of generalizing from their learning material and flexibility.
We say that policies are abstract ideas of what is allowed and what is off limits, from individual user preferences over administrative instruction documents to non-written social norms.
We define constraints as machine-executable representations of policies that have to be considered by the planner to produce plans that adhere to the constraints.
A symbolic constraint is a constraint in form of, e.g., a predicate-logical formula, as in classical access control.
A neural constraint is a constraint in form of a deep neural network.

Solving planning problems under constraints represents a classical topic.
But when scaling up the complexity of tasks, constraints, and environments to task universality in human-inhabited private spaces, classical methods might not deliver the performance required for real-world use like humanoid robots in assisted living scenarios.
In particular, there might be an unacceptable overhead in manual specification of constraints as well as an equally unacceptable overhead in computing costs.
As we will indicate in~\cref{sec:ac_tp}, however, neural planning struggles to actually ensure safety, security, and privacy constraints in its generated plans.
Constrained task-universal planning requires a hybrid method, for which we see large language models, due to their proficiency in a broad range of tasks and internalized “common sense”, and neuro-symbolic artificial intelligence, due to bridging the gap between localist and distributed data representation, as most promising research directions (c.f.~\cref{sec:modern}).

\section{Analysis of Standard Methods}
\label{sec:ac_tp}

\begin{figure}
    \centering
    \includegraphics[width=\linewidth]{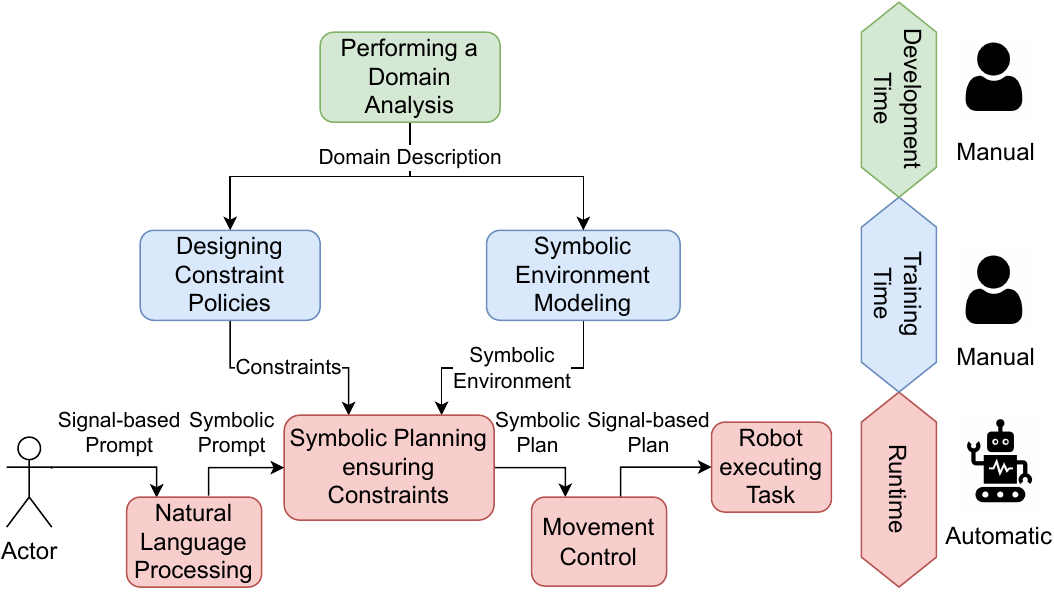}
    \caption{Workflow of designing a robot task planner using symbolic planning and symbolic constraints.
    We differentiate between three phases: In the development phase, the rules for the constraints are designed and the modeling for the symbolic planner is done.
    The training phase consists of the incorporation of such constraints into the planner. During runtime, the plan is generated and executed.}
    \label{fig:symbolic_rule_flow}
\end{figure}

\subsection{Constraint-Ensuring Symbolic Planning} 
\label{sec:symbolic-planning}
In symbolic planning, a planner finds a symbolic plan by searching through possible robot actions, trying to find transitions that transform the current state of the robot and its environment to the goal state inferred from the user prompt.
In practice, the de-facto standard for this is the Planning Domain Definition Language (PDDL)~\cite{aeronautiques1998pddl}, which we used in the following.
In our case, the initial state is the current state of the robot and its environment, the goal state is the goal that the prompt described as target, and the transition rules are the actions the robot is capable of.
The symbolic planner will then try to find a plan, i.e., a sequence of transitions that transforms the initial state into the goal state.
Each action has preconditions that need to hold in order for it to be executed.
Typical preconditions are that the robot has to be at a table and empty-handed to grab an item, or that an item has to be unobstructed to be grabbed.
Incorporating safety, security, and privacy constraints in form of preconditions is a natural approach to generate plans that are aware of such constraints, as shown in~\cref{lst:pddl}.

\begin{lstlisting}[caption=Excerpt from our PDDL domain description,label={lst:pddl},language=PDDL]
(:action clean_from_table
:parameters
    (?robot - robot
    ?table - location
    ?obj - on_table
    ?remove - location)
:precondition (and
    (non_personal ?obj)
    (at ?robot ?table)
    (at ?obj ?table)
    (remove_loc ?remove)
)
:effect (and
    (not (at ?obj ?table))
    (at ?obj ?remove)
))
...
(:init
    (at robot start)
    (at newspaper table)
    (at diary table)
    (at dishes table)
    (non_personal dishes)
    (non_personal newspaper)
    (personal diary)
    (remove_loc remove))
(:goal
    (forall (?obj - on_table) (or
    (and (non_personal ?obj) (at ?obj remove))
    (and (personal ?obj) (at ?obj table))
))))
\end{lstlisting}

Since the abstraction of a PDDL transition corresponds to ACAC's main abstraction of an activity, these two concepts can be easily combined.
This workflow is illustrated in \cref{fig:symbolic_rule_flow}.
We identify three different phases when designing such planner.
Only during runtime we automatically generate plans, while during development and training time, manual work needs to be done.
We were able to map\footnote{Full artifacts available: \url{https://github.com/kit-dsn/how-to-raise-a-robot-beyond-ac}\label{ft:arfitacts}} the preconditions of ACAC directly into the preconditions of PDDL, as well as the resulting conditions of an activity, which could be represented as \emph{effect} in PDDL.

However, ACAC's contextual conditions and current conditions, which are checked during the execution of an action, are not directly translatable.
Contextual conditions require PDDL extensions to query external data.
Current conditions require PDDL extensions that allow temporal planning and durative, interruptible actions.
While it is possible to incorporate toy examples of ACAC into PDDL, we question the scalability up to the humanoid robots in a care home use case without requiring an equally complex manual specification.
Every possible action, every possible object and every ACAC policy needs to be considered, inherently prohibiting such systems from being task-universal.
While symbolic artificial intelligence can learn incrementally in the field and can thereby reduce manual specification complexity while still being understandable and tunable by humans, adapting symbolic artificial intelligence to a broad range of situations is still a `very manual' process.

By relaxation from constraint-ensuring to constraint-observing planning, recent developments in the field of machine-learning based access control can be employed.
Feasibility of access decisions performed by machine learning methods such as neural networks was, e.g., shown by \citeauthor{nobi2022toward} \cite{nobi2022toward}.
However, current systems show quite large error margins, reaching a precision of about 0.9.
\citeauthor{nobi2022toward} propose the \emph{Deep Learning Based Access Control} model, where a neural network directly performs access decisions.
Such systems are mostly trained on access logs and further user and resource metadata.
To combine symbolic planning with a learned access control model, the planner queries the learned model in addition or instead of symbolic constraint formulas, and uses a learned heuristic function that classifies branches of the state space as dead ends if a constraint is violated.

\begin{figure}
    \centering
    \includegraphics[width=\linewidth]{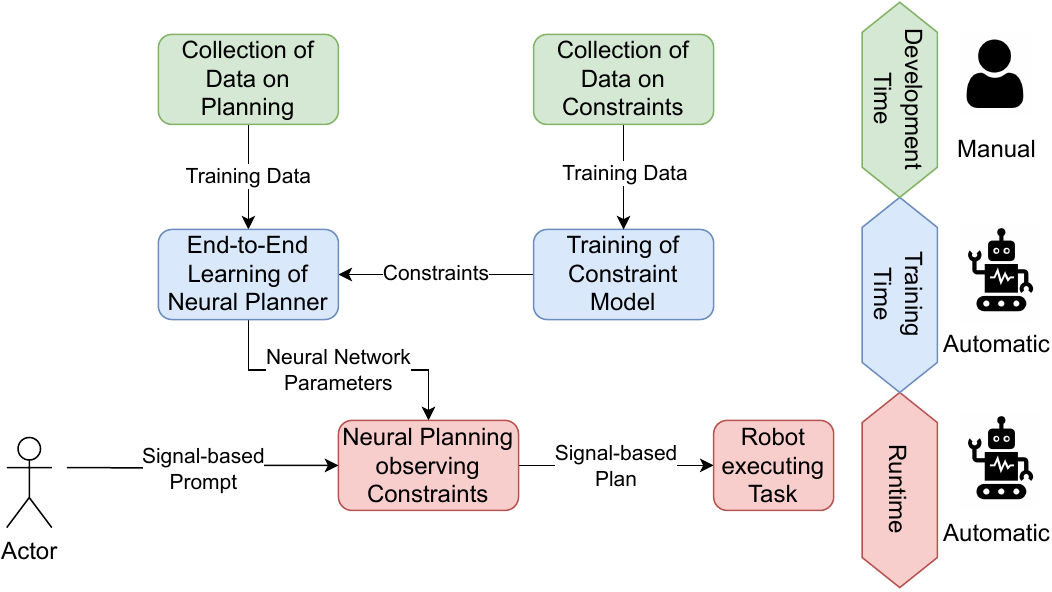}
    \caption{Workflow of designing a neural planner that satisfies neural constraints.
    The development phase now consists only of data collection.
    The training phase includes the actual training of the deep neural network, while the neural constraints are being fed into the training of the planner.
    During runtime, a forward pass through the network is performed.}
    \label{fig:flow_learned_learned}
\end{figure}

\subsection{Constraint-Observing Neural Planning} 
\label{sec:neural-planning}
Deep-learned neural planners promise to be scalable to complex deployments, and are the natural counterpart to deep-learning based access control.
When combining a neural constraint model with a neural planner, one gets a workflow as illustrated in \cref{fig:flow_learned_learned}.
Again we identify three different phases.
Compared to symbolic planning with symbolic constraints, i.e., classical access control, we can now perform the training phase automatically.
Only the development needs manual work, which improves scalability.

However, incremental learning without catastrophic forgetting is still an unsolved problem in deep neural networks.
Therefore, changes in the policy make retraining of the neural planner and neural constraints necessary.
Still, ways to tackle policy changes in a manageable way have been shown regarding DLBAC administration~\cite{nobi2022administration}.
The main idea is to retrain the networks in small steps, only using training data that represents the update that should be introduced.
This intends to only change the neural networks to incorporate the new data from the update, but this cannot be guaranteed.
One could also consider incorporating neural constraints into the reward calculation of reinforcement-learning based planners.

We can also combine neural planning with symbolic activity control.
Analogously to the previous paragraph, but unconventionally for Generative Adversarial Network (GANs), one might use symbolic constraints as discriminators to train the end-to-end learning-based neural planners.
It is challenging to provide a distributed representation of constraints, since symbolic constraints are localist representations.
As approximation, a first step can be realized by using the differentiable loss function provided by existing deep learning based constraint systems, which can be used to train the network.
Although this approach might not be as scalable to complex deployments due to manual policy engineering as learning neural constraints, it is more efficient to administrate since the symbolic constraints do not need to be retrained after a change.
For our running example, we now need to incorporate policies on what are personal items again, which makes the whole process inherently more complex.
Even though the neural planner learns its behavior, extensive policy engineering needs to be done.
Once the training of the end-to-end learning-based planner is finished, this approach behaves similarly to using a deep-learning based activity control system:
Given a prompt, the robot performs a forward pass through the neural planning network and receives a plan that probably observes the desired constraints.


\section{Analysis of Upcoming Methods} 
\label{sec:modern}
While symbolic and neural constraints are quite different compared to each other, the process of integrating any type of constraints in either symbolic or neural planning is similar.
Therefore, one is not limited to choosing only one combination.
We instead argue for a hybrid approach to combine the best of both worlds:
one can use neural planning and neural constraints for versatility and universality, but ensure critical policies as symbolic constraints during runtime.
For critical situations with limited universality, one can also fall back to symbolic planning to find an accurate solution slowly.
Neural systems could also serve as recommendation engine for new policies that enhance the existing logical solution.

Recently, Large Language Models (LLMs) became a popular method for artificial intelligence.
Especially the model ChatGPT by OpenAI \cite{openai2023gpt4} caught the attention of the general public, due to its seemingly close resemblance of human conversation while also being able to give symbolic answers, like solving a task in a programming language or even PDDL.
Due to their proficiency in processing and generating both signal- and symbol-based input and output, they are a natural building block for hybrid systems.

Neuro-symbolic artificial intelligence stands for the combination of symbolic and neural approaches to solve problems, and promises the “best of both worlds”, i.e., combining the advantages of both symbolic and neural methods.
Thus, the popularity of neuro-symbolic methods has risen in recent years \cite{sheth2023neurosymbolic}, and those methods promise to bring together the flexibility of neural planning with guaranteeing constraint satisfaction of symbolic planning.
We discuss LLMs using our toy example next, and then generalize to discuss neuro-symbolic methods to constraint-ensuring universal task planning in the following.

\subsection{LLMs as Planner and Knowledge Base}
\label{sec:llm}

\begin{figure}[htbp]
    \centering
    \includegraphics[width=\linewidth]{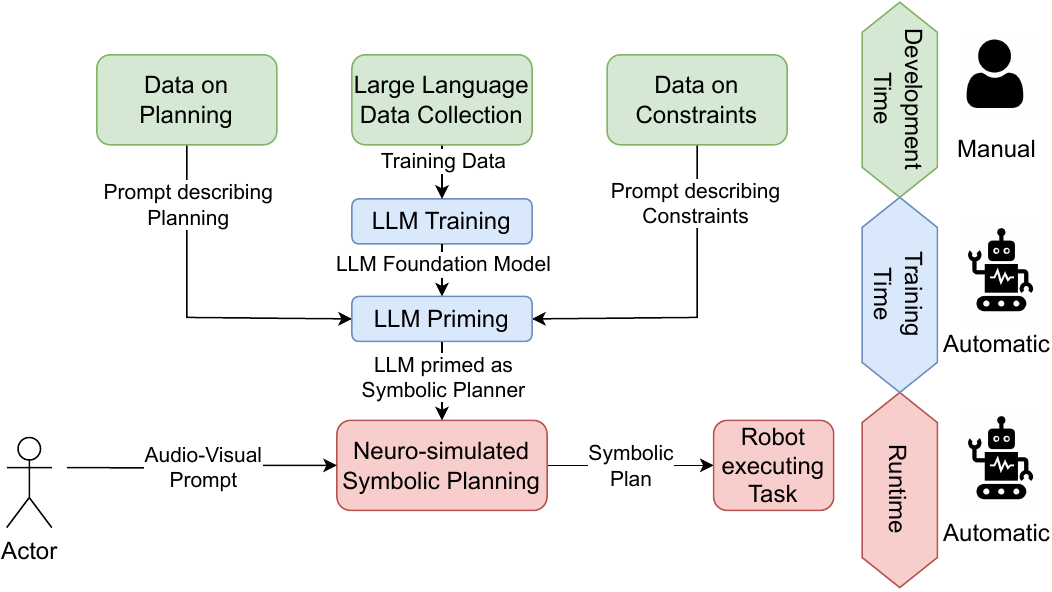}
    \caption{Workflow of designing a robot task planner using a multi-modal large language model as foundation model.
    In the development phase, the prompts for priming are manually engineered.}
    \label{fig:llm_flow}
\end{figure}
While LLMs represent a break-through for dialogues with humans, LLMs are capable of solving many kinds of problems given the right prompt.
Due to this versatility,  LLMs are foundational models, from which a problem-specific model is derived through priming.
The LLM can act on the priming because of the breadth and sheer size of its training data, which includes descriptions of planning problems, but also allows the models to generalize knowledge between domains -- for an LLM, solving a planning problem like the classical wolf, goat and cabbage river crossing problem in English is of no fundamental difference than solving it in another natural language, or in a formal planning language like PDDL.

We now describe an experiment with priming LLMs to behave like symbolic planers to solve constrained planing problems.
We can see such a workflow illustrated in \cref{fig:llm_flow}.
With recent advances in multimodal large language model, we envision the model to operate not only on textual information, but also on audio-visual user prompts.
We tested the primed LLM approach\footnotemark[1]{} by formulating our running example as natural-language prompt, and used ChatGPT 4 as constraint-observing neural planner to generate natural-language plans,
shown in \cref{lst:gptprompt}.
Specifically, lines 1 and 12 reduce ChatGPT to a constraint-observing neural planner by making it emulate a constraint-ensuring symbolic planner.
Lines 2-8 simulate sensors and memory, lines 9-11 instruct ChatGPT with a goal and constraints.
We obtain a plan that reaches the goal of cleaning the table, while not interacting with the most personal item on the table, the diary.
This experiment shows that we can use a GPT-based LLM to do neural planning while observing neural constraints inferred from natural-language prompts.

\begin{lstlisting}[float,caption={A prompt to simulate a symbolic planner for a basic domain description, and the ChatGPT 4 answer. Shared chat transcript at: \\ {\footnotesize \url{https://chat.openai.com/share/77d31c8a-239c-4d57-a96c-f70078f88503}}}, label={lst:gptprompt},keywords={},columns=fullflexible]
Suppose you are a symbolic planning system that generates plans for task-universal assistive and humanoid robots based on a domain description that is provided below.
Domain:
- Care home that uses robots to care for their residents. The care home has lots of personal rooms for the residents, common spaces like a dining area, and administrative spaces like a kitchen or offices
- Personal room of a resident, that has a table inside
- on the table are the following things:
	- newspaper that was paid for by the resident
	- dirty dishes that are owned by the care home and are cleaned by the care home
	- a diary that most certainly contains very private information
Goal: Remove as many objects from the table as possible
Constraints: Do not interact with objects that you would consider personal.
Possible actions: Move, analyze an object, take an object, place an object
Put out the plan in terms of a sequence of actions the robot performs. Put out the action you used and the object you interacted with or where the object was placed. You can only use the actions from the list above. Put out a list of actions and objects that the plan consists of. Summarize the reasoning for the overall plan as a single text paragraph. The plan must be conform with all given constraints.
# Exemplary output from ChatGPT 4
Plan Sequence:
1. Analyze object: newspaper
2. Analyze object: dirty dishes
3. Analyze object: diary
4. Take object: dirty dishes
5. Move to: kitchen or designated cleaning area
6. Place object: dirty dishes in cleaning area or dishwashing station
Reasoning:
Given the constraints and the goal, the robot needs to first distinguish between objects that are personal and objects that belong to the care home. Analyzing each object ensures the robot can determine its nature and importance. The newspaper, although bought by the resident, could be seen as potentially non-intrusive. However, it was paid for by the resident which could imply some personal value, so it's left untouched. The diary is most certainly private, so it's not touched. The dirty dishes belong to the care home and must be cleaned, so they are the only objects removed from the table, in line with the constraints set.
\end{lstlisting}

One interesting effect of using LLMs for constraint-observing planning is their performance on problems that require knowledge of \emph{common sense}.
Our toy example delivered an insight into such behavior.
We tasked the LLM-based planner with only handling objects that are not considered private.
From experience and intuition, humans know that a diary is a very private object and dirty dishes are not, but machines need to encode such information somehow.
When using LLMs as planner, such information does not need to be given explicitly to the planner, because the LLM can infer it from its training data.
Since it is trained on extensive textual data that encodes a lot of human interaction, such social norms are known.
When asked whether a diary is private, a \emph{yes} has a higher probability than a \emph{no}.
The LLM therefore not only acts as a symbolic planner, but also as knowledge base for common sense and object attribution for the simulated planner.
This integrated knowledge is of tremendous value for systems deployed in real world applications, because such social norms will no longer need to be encoded directly, if a system should obey to them.
Especially humanoid assistive robots act in very sensitive environments and will need to follow a very complex set of rules.
While some of these rules will be available in the form of certain documents, some rules will only exist in the form of social norms, for which no single “ISO norm” document exists.
LLMs provide a solution to encoding such norms, without extensive work on building a symbolic version of them.
Thus, LLMs serve as knowledge bases for such systems.

\begin{lstlisting}[float,caption={A river crossing puzzle as stated in Wikipedia~\cite{wiki-river-crossing-puzzle}, but with the wolf replaced by a pig.
The ChatGPT 4 solution leaves either of the herbivores alone with the cabbage. Shared chat transcript: {\footnotesize \url{https://chat.openai.com/share/b9eef797-a4ad-4b50-a9d1-3cc294e744f6}
}}, label={lst:gpt-pig-goat-cabbage},keywords={},columns=fullflexible]
A farmer with a pig, a goat, and a cabbage must cross a river by boat. The boat can carry only the farmer and a single item. If left unattended together, the goat or the pig would eat the cabbage. How can they cross the river without anything being eaten?
# Exemplary output from ChatGPT 4
The puzzle you've described is a classic river crossing problem, though it's typically presented with a wolf, a goat, and a cabbage. The solution to this problem, where the pig or the goat would eat the cabbage if left unattended, is as follows:
1. The farmer takes the goat (or pig, it doesn't matter which) across the river first and leaves it on the other side.
2. The farmer then goes back alone to the original side.
3. He takes the cabbage across the river next.
4. After leaving the cabbage on the other side, he takes the goat (or pig) that was left there during the first trip back with him to the original side.
5. He leaves the goat (or pig) on the original side and takes the pig (or goat, the one he hasn't taken across yet) to the other side.
6. The farmer leaves the pig (or goat) with the cabbage on the other side and goes back to the original side to get the goat (or pig).
7. Finally, he takes the goat (or pig) across the river.
Now, all three items - the pig, the goat, and the cabbage - are safely on the other side of the river. At no point were the cabbage and the animals left unattended together, so nothing was eaten.
\end{lstlisting}

In our experiments, the dishes were always moved to the kitchen, while the newspaper and diary were always considered too personal to interact with.
However, with ChatGPT 3.5, the newspaper was sometimes interacted with and sometimes not, showing probabilistic instabilities not present with a (non-probabilistic) symbolic planner.
These instabilities stem from the fact that the LLM only simulated a symbolic planner, using neural methods.
The LLM chose among the predicted text continuations with the highest probability, and if two alternatives have no clear winner, the solution will vary.
In contrast to actual symbolic planning, there are no guarantees on stability, correctness, or other properties of the solution.
In contrast, systems built to perform robot task planning usually have guarantees regarding the validity and correctness of produced plans, but LLMs instructed to behave like symbolic planners cannot provide guarantees on stability, correctness, or other properties of the solution.
Planning-primed LLMs produce outputs that \emph{sound like} a plan, but that are not necessarily a correct or even valid plan.
For example, \cref{lst:gpt-pig-goat-cabbage} shows a prompt for a variation of the river-crossing problem usually formulated using a wolf, a goat, and a cabbage, which is a common way to test the limits of LLMs (c.f., e.g., Gerhard Paaß in~\cite{ct-chatbots-reinlegen}).
Here, the wolf is replaced with a pig, and the constraints are stated accordingly that neither of the animals can be left unattended with the cabbage.
While ChatGPT 4 confirms this fact, its plan does not ensure this constraint, as it leaves one of the animals alone with the cabbage, due to missing symbolic understanding of the problem.
Instead, it provides a solution similar to the regular solution with the wolf, due to its prevalence in the training data.
This example demonstrates that while likelihood-based text completion can simulate symbolic reasoning, it is currently no replacement for it.

While LLMs worked well for showcasing versatile planning behavior based on the knowledge of social norms in our toy example, the river-crossing problem example confirms the need for symbolic reasoning to get correct solutions.
The emerging field of neuro-symbolic artificial intelligence acts as a promising direction to combine versatility and correctness, which we look into next.

\subsection{Integration of Neural and Symbolic AI}

The term \emph{neuro-symbolic AI} directly describes systems that combine symbolic with neural approaches.
The above-mentioned hybrid systems would fall into this category.
Neuro-symbolic AI systems are differentiated on the extent of the integration of symbolic data representation in the neural process. Kautz~\cite{kautz22thirdaisummer} differentiates between six different types of neuro-symbolic integration (c.f.~\cref{tab:datasets}).
This classification ranges from standard deep learning or large language models, where the input and output can be symbolic, but the inner workings are not, to type six integration, where full symbolic reasoning should be possible within the neural engine. We especially want to emphasize how the integration of symbolic reasoning into the training process happens on each level.
For our analysis, we consider each level and give an example mapping for our constrained task planning scenario.

\renewcommand{\arraystretch}{1.5}
\newcommand{\lvl}[1]{{\ttfamily\bfseries #1}}

\begin{table}[t]
      \centering
      \begin{tabular}{ r p{0.4\linewidth} p{0.45\linewidth} }
        \textbf{\raggedleft Lvl} & \textbf{Description based on Garcez et al.\cite{garcez2023neurosymbolic}} & \textbf{Example for Constrained Robot Task Planning} \\
        \hline
        \lvl{1} & Standard neural networks deep-learned for using symbolic input and output & Learned neural network planner that produces symbolic plans from symbolic prompts \\
        \lvl{2} & A neural network is loosely coupled with a symbolic problem solver & A LLM is used as knowledge base for a symbolic constraint system \\
        \lvl{3} & A neural network focusing on one task is interacting with a symbolic system performing a complementary task & A neural planner produces symbolic plans, which are verified post-planning to ensure symbolic security and privacy constraints trough a symbolic constraint system \\
        \lvl{4} & Symbolic knowledge is given into the training set of a neural network & The training data is generated by a simulator based on symbolic domain and constraint representations \\
        \lvl{5} & Symbolic logic is mapped onto an embedding which acts as a soft-constraint on the network's loss function & The loss function of a neural planner has an embedded symbolic constraint system, penalizing illegal plans and emphasizing legal ones \\
        \lvl{6} & True symbolic reasoning inside a neural engine & A neural planner runs completely on a symbolic representation of the domain and its constraints \\
      \end{tabular}
      \caption{Levels of neuro-symbolic integration according to Kautz \cite{kautz22thirdaisummer} and their respective realization for robot-task planning.}
      \label{tab:datasets}
\end{table}

Our first experiment using a symbolic planner and PDDL from \cref{sec:symbolic-planning} has no neural reasoning at all, i.e., provides no neuro-symbolic integration, and therefore one could say it is a \textbf{“level zero”} system.
The same is true for our end-to-end learned neural planner architecture from \cref{sec:neural-planning}, as no symbolic representation is involved.

Kautz' categorization starts with \textbf{level one}, which is classical deep learning trained on symbolic input to produce symbolic output.
Since there is no symbolic reasoning inside the system, the data representation is purely distributed inside the neural engine.
The integration of constraint satisfaction may be possible by only using valid plans as training data, hoping to not generate invalid ones.
The LLM system from \cref{sec:llm} would fall in this category, if we prompted it with symbolic inputs to produce symbolic outputs, e.g. in PDDL.

On \textbf{level two}, there are loosely coupled systems.
This loose coupling is given if a symbolic problem solver uses a neural system as some kind of sub-routine, as stated by Kautz~\cite{kautz22thirdaisummer}.
In our constrained task planning case, this level could be reached with a symbolic planer like in our first experiment, enhanced to use a LLM as knowledge base.
The sub-routine would allow the symbolic planner to check an item for its level of privacy and the societal knowledge encoded in the LLM.
It is important to note that the training process of both parts is completely decoupled, i.e., the symbolic part has no influence on the training of the neural part and vice-versa.

The \textbf{third level} is, e.g., the “do not cross the red line” approach.
We again have a distinct neural and symbolic system.
On this level, the output of one is given as input to the other one.
While on level two, the data flow was primarily handled by one of the systems, with the other one being just a sub-routine, it is now handled by both systems.
A typical example is the pipeline of a neural computer vision system which produces the symbolic input for a symbolic planner, whose symbolic output is translated to motor signals by another neural system.
In our scenario, an example would be a pipeline that feeds the symbolic plans of an LLM-based planner to a symbolic constraint enforcement system, as seen with Yang et al.'s LLM coupled with a linear temporal logic constraint enforcement~\cite{yang2023plug}.
An important distinction of level three in contrast to level two is that now both systems take part in a joint training procedure.
Kautz mentions a feedback loop that flows back from the symbolic system to optimize the training of the neural system.

\textbf{Level four} represents systems where symbolic knowledge is put into the neural system by encoding it into synthetic training data.
This approach essentially means to use symbolic knowledge on the problem domain, but also general symbolic knowledge of physics or mathematics applied to the domain, to generate valid and invalid solutions to many problem instances.
Note that finding a problem instance for a given, randomly-generated solution, i.e., to generate a domain in which a given plan does not violate any constraints, might be much faster than the other way around, i.e., actual symbolic planning.
An exemplary system on this level would be a neural constraint enforcement machine trained on synthetic decisions logs.
The training data would consist of access requests and corresponding access decisions, generated from a set of symbolic constraints that formalize the access policy.
Analogously, a neural task planner could be trained with synthetic pairs of tasks and suitable plans.
It is important to note the difference to a training on real world access logs here:
All synthetic training samples have symbolic knowledge inherently encoded.
This is not necessarily the case on real world data, where the inherent logic might be overlaid by pertubations, or not be present at all.
Symbolically generated, synthetic training data further deepens the integration of the symbolic reasoning into the training process of the neural system.
On level three, it was a feedback loop back into the training, now it is directly embedded.

On \textbf{level five}, one looks at systems where symbolic logic is directly mapped onto the loss function of the neural system.
This level is especially challenging because the distributed data representation of a symbolic system is typically not differentiable.
In our planning example, this approach would mean that for example a symbolic constraint enforcement is directly mapped into the loss function of the neural planner.
Plans that are valid would be emphasized by it and plans that are not valid would be penalized.
Since the symbolic logic is directly in the loss function of the neural system, we cannot distinguish the symbolic part in the training process anymore, it has become part of it.

The last step is further strengthening this integration of symbolic logic into the neural system.
Such systems are described by \textbf{level six} neuro-symbolic systems, representing the deepest neuro-symbolic integration.
On this level, a real distinction of neural and symbolic parts is almost impossible: 
The symbolic knowledge is deeply encoded in the whole neural system.
Systems on this level might be compared to human brains, bridging the gap between automatic and instinctive part of the brain and the slow and logical part, according to Kahnemanns notion of Systems 1 and 2~\cite{kahnemann2011thinking}.
State of the art neural systems correspond to the System 1 fast and instinctive part, and are not able to bridge the gap to also think slow and logical.
This behavior is exemplarily demonstrated in \cref{lst:gpt-pig-goat-cabbage}, where ChatGPT “reads over” the peculiarities of the specific problem instance, and instead provides a solution similar to the one for the common instance presumably more prevalent in its training data.
Our example for this level is a neural planner that runs completely on a symbolic representation and its constraints.
The vagueness of this example is because such system do not exist currently, and it is still unclear how they could look like in practice, or whether they are feasible at all.
However, such level six systems would be learned symbolic reasoning machines.
Due to their capability of “thinking, fast and slow” both in the neural and symbolic world, they would be a game-changer for constrained task planning.
One would be able to completely explain the machines and guarantee their behavior, neither of which is currently possible using black-box deep-learned neural methods. 



\section{Challenges Ahead}

\subsection{Neuro-Symbolic Integration of Constraint Satisfaction and Task Planning}


On current systems, we observe that task planning for real-world robots scales best to complex situations using neural planners, while in the access control community, symbolic approaches are most prevalent to strictly enforce security and privacy constraints.
Therefore, we see the challenge of neuro-symbolic integration for constrained robot task planning in integrating a neural planner with a symbolic constraint system.
While deepening the integration of the symbolic and neural spheres is still an open research problem,
we present goals and challenges for deep neuro-symbolic integration for constrained robot task planning in the remainder of this section.

Deep-learned neural planning requires a distributed representation of plans instead of a localist representation as a list of task actions.
With end-to-end learning-based robot task planners, we get this distributed representation of plans for free.
With the feature space being the sensor data, we can directly perform neural operations on plan-level, which is encoded by the internal weights of the neural network that represents the planner.
To interact with a neural planner, constraints need a distributed representation as well.
Recent developments in deep learning based access control promise comparable approaches to the distributed representation of end-to-end learning based robot planners.
The deep neural networks that perform access decisions encode distributed constraints in their internal weights, and provide a feature space on which such constraints can be learned.
However, satisfaction of constraints in distributed representation can still not be guaranteed, as seen in the sections above.

Thus the integration of constraint-ensuring neural task planning and comes down to the following challenges that can be considered variations of the well-known signal-to-symbol gap challenges:
\begin{itemize}
    \item How can a distributed representation of planning work together with a distributed representation of constraints?
    \item How can symbolic algorithms reason on a suitable distributed data representations?
    \item How to learn symbolic knowledge on distributed feature spaces?
\end{itemize}

In this paper, we described approaches up to level two in the previous sections.
As a research community, we can currently build neuro-symbolic systems up to level three and four.
For level three, we can use existing systems and plug them together,
level four can be realized using synthetic training data generated from symbolic knowledge on planning and constraints.
Obviously, the leap to level five is the challenges, however, it would also bring humanoid assistive robots on a new level that might be necessary for practical task-universal usage in the real world.
Existing neural constraint enforcement systems such as DLBAC~\cite{nobi2022toward} show that the integration of such systems into the loss function of planners appears on the horizon of current research.
Controlled text generation of LLMs is also a current research topic, addressing the question of how to make a LLM produce text of a specific kind, as seen with Dathari et al.~\cite{dathathri2019plug} or Keskar et al.~\cite{keskar2019ctrl}.
Such imposing of constraints are a promising fit for imposing constraints on planners to use LLMs in some form, as knowledge base or even directly as planner.
While these controllers also are neural networks as of today and therefore are level four systems at most, an approach with symbolic controllers would be a level five neuro-symbolic AI.

A type six neuro-symbolic system that utilizes truly symbolic reasoning inside a neural engine would make many problems of constraining task-universal robots disappear, and thereby represent a major break-through for secure, private, but task-universal humanoid assistive robots in the real world.
Such a highly-integrated neuro-symbolic system would be as task-universal, flexible and fast as today's neural planners, but still able to guarantee that learned constraints would be met as today's symbolic planners.
A truly neuro-symbolic system would easily be able to scale in the dimension of task and constraint complexity, because those could be learned and would not need to be manually designed.
However, while we seem to have major parts for functioning level five neuro-symbolic systems, level six still presents a grand challenge that is possibly much farther off in the future, if achievable at all.

However, a deep level six neuro-symbolic integration might not be necessary for acceptable trade-offs regarding constrained task planning.
Pure symbolic problem solving always comes with the advantage of guaranteeing certain behavior, which neural approaches cannot provide.
For example, one could think of high-level symbolic planners that provide a general idea, and only fine-grained planning is done using neural systems.
The same idea can be used for constraints that are imposed or observed by such planning systems.
The integration of symbolic reasoning might also be done by imposing symbolic \emph{red lines} to the neural planning system.
The neural planner would then be constrained by a symbolic system that encodes rules of high importance.
Aspects with a lower security level can however be constrained by a neural planning system, because errors do not have such an impact there.

\subsection{Next Steps}
In this paper, we argued that neuro-symbolic hybrid approaches are necessary to bring constrained task-universal planning to real-world problem complexity.
We listed many variations of possible hybrid approaches, which still need to prove their feasibility in real robot experiments.
Feasibility especially means how well the different approaches cope with practical task and policy complexity in e.g. our assistive humanoid robot in care homes scenario.
A practical feasibility analysis could also determine necessary and sufficient levels of neuro-symbolic integration for the scenario, finding out what type of neuro-symbolic integration would fit best.
While deploying neuro-symbolic systems on real robots might be an extensive challenge, simulating such situations with the necessary degree of realism using information gathered from care homes seems feasible.

In the space of possible neuro-symbolic approaches, we see LLMs as most promising building block due to their inherent knowledge on social norms and ability to work with multi-modal, layered instructions - we imagine a LLM that is primed to behave like a humanoid assistive robot in a care home using documents and norms for human trainees, that use input from the robot's camera, other sensors and memory to perform a task given as voice command in natural language.
However, further research into guaranteeing valid plans or even correct plans when using LLMs is necessary -- we see a promising direction in building upon advances in controlling LLM output as a way to make LLM-based planners adhere to constraints.
In addition, performance of LLM-based planning in comparison to standard symbolic and neural planning systems in terms of speed and correctness is yet to be evaluated.
Specific to LLMs used as knowledge base for planners, the general performance of inferring social norms also has to be evaluated.

Orthogonal to testing the feasibility of current ideas for hybrid approaches, neuro-symbolic AI advances as a field of itself.
Any progress in this field, especially in the integration of a neural engine and a symbolic constraint system, would be directly applicable in constrained task planning for humanoid assistive robots.

\section{Conclusion}
\label{sec:conclusion}
We discussed combinations of symbolic and neural constrained task planning approaches.
We highlighted their trade-offs and showed their benefits and shortcomings.
While neural planners scale better to complex deployments in terms of required manual specification, they introduce an error margin and hinder administrability.
The usage of neural systems such as LLMs as knowledge base for access decisions presents a direct way of incorporating social norms into such access decisions.
Designing symbolic planning systems, however, requires more manual tasks than neural planning systems.
While neural systems seem have in advantage when scaling problem complexity up to task universality, to actually teach them security, they do not only challenge us to quantify what we mean with security, but also to make that notion differentiable.
In order to create deployable systems one should, therefore, strive for hybrid designs to combine best of both worlds: neural planning with neural constraints leads to probable observance of policies, critical policies will be safeguarded by symbolic constraints.
Neuro-symbolic AI promises to advance this field of constrained task planning of assistive humanoid robots.

\begin{acks}
We like to thank Jan Niehues for his valuable comments, in particular for his hints on controlled text generation and controller architectures. This work has been supported by the project ``Stay young with robots'' (JuBot).
The JuBot project was made possible by funding from the Carl Zeiss Foundation.
\end{acks}

\bibliographystyle{ACM-Reference-Format}
\bibliography{main}


\begin{thebibliography}{16}


\ifx \showCODEN    \undefined \def \showCODEN     #1{\unskip}     \fi
\ifx \showDOI      \undefined \def \showDOI       #1{#1}\fi
\ifx \showISBNx    \undefined \def \showISBNx     #1{\unskip}     \fi
\ifx \showISBNxiii \undefined \def \showISBNxiii  #1{\unskip}     \fi
\ifx \showISSN     \undefined \def \showISSN      #1{\unskip}     \fi
\ifx \showLCCN     \undefined \def \showLCCN      #1{\unskip}     \fi
\ifx \shownote     \undefined \def \shownote      #1{#1}          \fi
\ifx \showarticletitle \undefined \def \showarticletitle #1{#1}   \fi
\ifx \showURL      \undefined \def \showURL       {\relax}        \fi
\providecommand\bibfield[2]{#2}
\providecommand\bibinfo[2]{#2}
\providecommand\natexlab[1]{#1}
\providecommand\showeprint[2][]{arXiv:#2}

\bibitem[Bayreuther et~al\mbox{.}(2022)]%
        {bayreuther2022bluesky}
\bibfield{author}{\bibinfo{person}{Saskia Bayreuther}, \bibinfo{person}{Florian
  Jacob}, \bibinfo{person}{Markus Grotz}, \bibinfo{person}{Rainer Kartmann},
  \bibinfo{person}{Fabian Peller-Konrad}, \bibinfo{person}{Fabian Paus},
  \bibinfo{person}{Hannes Hartenstein}, {and} \bibinfo{person}{Tamim Asfour}.}
  \bibinfo{year}{2022}\natexlab{}.
\newblock \showarticletitle{BlueSky: Combining Task Planning and
  Activity-Centric Access Control for Assistive Humanoid Robots}. In
  \bibinfo{booktitle}{\emph{Proceedings of the 27th ACM Symposium on Access
  Control Models and Technologies}} \emph{(\bibinfo{series}{SACMAT '22})}.
  \bibinfo{publisher}{ACM}, \bibinfo{address}{New York},
  \bibinfo{pages}{185–194}.
\newblock
\showISBNx{9781450393577}
\urldef\tempurl%
\url{https://doi.org/10.1145/3532105.3535018}
\showDOI{\tempurl}


\bibitem[Dathathri et~al\mbox{.}(2020)]%
        {dathathri2019plug}
\bibfield{author}{\bibinfo{person}{Sumanth Dathathri}, \bibinfo{person}{Andrea
  Madotto}, \bibinfo{person}{Janice Lan}, \bibinfo{person}{Jane Hung},
  \bibinfo{person}{Eric Frank}, \bibinfo{person}{Piero Molino},
  \bibinfo{person}{Jason Yosinski}, {and} \bibinfo{person}{Rosanne Liu}.}
  \bibinfo{year}{2020}\natexlab{}.
\newblock \bibinfo{title}{Plug and Play Language Models: A Simple Approach to
  Controlled Text Generation}.
\newblock
\newblock
\showeprint[arxiv]{1912.02164}~[cs.CL]


\bibitem[Garcez and Lamb(2023)]%
        {garcez2023neurosymbolic}
\bibfield{author}{\bibinfo{person}{Artur~d’Avila Garcez} {and}
  \bibinfo{person}{Luis~C. Lamb}.} \bibinfo{year}{2023}\natexlab{}.
\newblock \showarticletitle{Neurosymbolic AI: The 3rd wave}.
\newblock \bibinfo{journal}{\emph{Artificial Intelligence Review}}
  (\bibinfo{year}{2023}).
\newblock
\urldef\tempurl%
\url{https://doi.org/10.1007/s10462-023-10448-w}
\showDOI{\tempurl}


\bibitem[Grävemeyer(2023)]%
        {ct-chatbots-reinlegen}
\bibfield{author}{\bibinfo{person}{Arne Grävemeyer}.}
  \bibinfo{year}{2023}\natexlab{}.
\newblock \showarticletitle{Chatbots reinlegen – Wie man KI-Sprachgeneratoren
  entlarvt}.
\newblock \bibinfo{journal}{\emph{c't}}  \bibinfo{volume}{16}
  (\bibinfo{year}{2023}), \bibinfo{pages}{116--119}.
\newblock
\urldef\tempurl%
\url{https://www.heise.de/select/ct/2023/16/2313010292321745772}
\showURL{%
\tempurl}
\newblock
\shownote{(“Fooling Chatbots - How to Expose AI Speech Generators.”)}.


\bibitem[Gupta and Sandhu(2021)]%
        {gupta2021towards}
\bibfield{author}{\bibinfo{person}{Maanak Gupta} {and} \bibinfo{person}{Ravi
  Sandhu}.} \bibinfo{year}{2021}\natexlab{}.
\newblock \showarticletitle{Towards Activity-Centric Access Control for Smart
  Collaborative Ecosystems}. In \bibinfo{booktitle}{\emph{Proceedings of the
  26th ACM Symposium on Access Control Models and Technologies}}
  \emph{(\bibinfo{series}{SACMAT '21})}. \bibinfo{publisher}{ACM},
  \bibinfo{address}{New York}, \bibinfo{pages}{155–164}.
\newblock
\showISBNx{9781450383653}
\urldef\tempurl%
\url{https://doi.org/10.1145/3450569.3463559}
\showDOI{\tempurl}


\bibitem[Hemken et~al\mbox{.}(2023)]%
        {how-to-raise-a-robot-poster}
\bibfield{author}{\bibinfo{person}{Niklas Hemken}, \bibinfo{person}{Florian
  Jacob}, \bibinfo{person}{Fabian Peller-Konrad}, \bibinfo{person}{Rainer
  Kartmann}, \bibinfo{person}{Tamim Asfour}, {and} \bibinfo{person}{Hannes
  Hartenstein}.} \bibinfo{year}{2023}\natexlab{}.
\newblock \showarticletitle{Poster: How to Raise a Robot - Beyond Access
  Control Constraints in Assistive Humanoid Robots}. In
  \bibinfo{booktitle}{\emph{Proceedings of the 28th ACM Symposium on Access
  Control Models and Technologies}} (Trento, Italy)
  \emph{(\bibinfo{series}{SACMAT '23})}. \bibinfo{publisher}{Association for
  Computing Machinery}, \bibinfo{address}{New York, NY, USA},
  \bibinfo{pages}{55–57}.
\newblock
\showISBNx{9798400701733}
\urldef\tempurl%
\url{https://doi.org/10.1145/3589608.3595078}
\showDOI{\tempurl}


\bibitem[Howe et~al\mbox{.}(1998)]%
        {aeronautiques1998pddl}
\bibfield{author}{\bibinfo{person}{Adele Howe}, \bibinfo{person}{Craig
  Knoblock}, \bibinfo{person}{Drew McDermott}, \bibinfo{person}{Ashwin Ram},
  \bibinfo{person}{Manuela Veloso}, \bibinfo{person}{Daniel Weld}, {and}
  \bibinfo{person}{David Wilkins}.} \bibinfo{year}{1998}\natexlab{}.
\newblock \bibinfo{booktitle}{\emph{{PDDL} – The Planning Domain Definition
  Language}}.
\newblock \bibinfo{type}{{T}echnical {R}eport}. \bibinfo{institution}{Yale
  Center for Computational Vision and Control}.
\newblock


\bibitem[Kahneman(2011)]%
        {kahnemann2011thinking}
\bibfield{author}{\bibinfo{person}{Daniel Kahneman}.}
  \bibinfo{year}{2011}\natexlab{}.
\newblock \bibinfo{booktitle}{\emph{Thinking, Fast and Slow}}.
\newblock \bibinfo{publisher}{Farrar, Straus and Giroux}, \bibinfo{address}{New
  York}.
\newblock
\showISBNx{978-0374275631}


\bibitem[Kautz(2022)]%
        {kautz22thirdaisummer}
\bibfield{author}{\bibinfo{person}{Henry~A. Kautz}.}
  \bibinfo{year}{2022}\natexlab{}.
\newblock \showarticletitle{The third AI summer: AAAI Robert S. Engelmore
  Memorial Lecture}.
\newblock \bibinfo{journal}{\emph{AI Magazine}} \bibinfo{volume}{43},
  \bibinfo{number}{1} (\bibinfo{year}{2022}), \bibinfo{pages}{105--125}.
\newblock
\urldef\tempurl%
\url{https://doi.org/10.1002/aaai.12036}
\showDOI{\tempurl}


\bibitem[Keskar et~al\mbox{.}(2019)]%
        {keskar2019ctrl}
\bibfield{author}{\bibinfo{person}{Nitish~Shirish Keskar},
  \bibinfo{person}{Bryan McCann}, \bibinfo{person}{Lav~R. Varshney},
  \bibinfo{person}{Caiming Xiong}, {and} \bibinfo{person}{Richard Socher}.}
  \bibinfo{year}{2019}\natexlab{}.
\newblock \bibinfo{title}{CTRL: A Conditional Transformer Language Model for
  Controllable Generation}.
\newblock
\newblock
\showeprint[arxiv]{1909.05858}~[cs.CL]


\bibitem[Nobi et~al\mbox{.}(2022a)]%
        {nobi2022administration}
\bibfield{author}{\bibinfo{person}{Mohammad~Nur Nobi}, \bibinfo{person}{Ram
  Krishnan}, \bibinfo{person}{Yufei Huang}, {and} \bibinfo{person}{Ravi
  Sandhu}.} \bibinfo{year}{2022}\natexlab{a}.
\newblock \showarticletitle{Administration of Machine Learning Based Access
  Control}. In \bibinfo{booktitle}{\emph{ESORICS 2022}}.
  \bibinfo{publisher}{Springer}, \bibinfo{address}{Cham},
  \bibinfo{pages}{189--210}.
\newblock
\urldef\tempurl%
\url{https://doi.org/10.1007/978-3-031-17146-8_10}
\showDOI{\tempurl}


\bibitem[Nobi et~al\mbox{.}(2022b)]%
        {nobi2022toward}
\bibfield{author}{\bibinfo{person}{Mohammad~Nur Nobi}, \bibinfo{person}{Ram
  Krishnan}, \bibinfo{person}{Yufei Huang}, \bibinfo{person}{Mehrnoosh
  Shakarami}, {and} \bibinfo{person}{Ravi Sandhu}.}
  \bibinfo{year}{2022}\natexlab{b}.
\newblock \showarticletitle{Toward Deep Learning Based Access Control}. In
  \bibinfo{booktitle}{\emph{Proceedings of the Twelfth ACM Conference on Data
  and Application Security and Privacy}} \emph{(\bibinfo{series}{CODASPY
  '22})}. \bibinfo{publisher}{ACM}, \bibinfo{address}{New York},
  \bibinfo{pages}{143–154}.
\newblock
\showISBNx{9781450392204}
\urldef\tempurl%
\url{https://doi.org/10.1145/3508398.3511497}
\showDOI{\tempurl}


\bibitem[OpenAI(2023)]%
        {openai2023gpt4}
\bibfield{author}{\bibinfo{person}{OpenAI}.} \bibinfo{year}{2023}\natexlab{}.
\newblock \bibinfo{title}{GPT-4 Technical Report}.
\newblock
\newblock
\showeprint[arxiv]{2303.08774}~[cs.CL]


\bibitem[Sheth et~al\mbox{.}(2023)]%
        {sheth2023neurosymbolic}
\bibfield{author}{\bibinfo{person}{Amit Sheth}, \bibinfo{person}{Kaushik Roy},
  {and} \bibinfo{person}{Manas Gaur}.} \bibinfo{year}{2023}\natexlab{}.
\newblock \bibinfo{title}{Neurosymbolic AI -- Why, What, and How}.
\newblock
\newblock
\showeprint[arxiv]{2305.00813}~[cs.AI]


\bibitem[{Wikipedia contributors}(2023)]%
        {wiki-river-crossing-puzzle}
\bibfield{author}{\bibinfo{person}{{Wikipedia contributors}}.}
  \bibinfo{year}{2023}\natexlab{}.
\newblock \bibinfo{booktitle}{\emph{River crossing puzzle}}.
\newblock
\urldef\tempurl%
\url{https://en.wikipedia.org/w/index.php?title=River_crossing_puzzle&oldid=1150277959}
\showURL{%
\tempurl}


\bibitem[Yang et~al\mbox{.}(2023)]%
        {yang2023plug}
\bibfield{author}{\bibinfo{person}{Ziyi Yang}, \bibinfo{person}{Shreyas~S.
  Raman}, \bibinfo{person}{Ankit Shah}, {and} \bibinfo{person}{Stefanie
  Tellex}.} \bibinfo{year}{2023}\natexlab{}.
\newblock \bibinfo{title}{Plug in the Safety Chip: Enforcing Constraints for
  LLM-driven Robot Agents}.
\newblock
\newblock
\showeprint[arxiv]{2309.09919}~[cs.RO]


\end{thebibliography}

\end{document}